\documentclass[sigconf,screen,nonacm]{acmart}

\usepackage{microtype}
\usepackage{graphicx}
\usepackage{subcaption}
\usepackage{booktabs} % for professional tables
\usepackage{wrapfig}
\usepackage{multicol}
\usepackage[T1]{fontenc}
\usepackage[utf8]{inputenc}
\usepackage{listings}
\usepackage{xcolor}
\usepackage{xspace}
\usepackage{hyperref}
\usepackage{booktabs}   % \toprule \midrule \bottomrule
\usepackage{tabularx}   % tabularx 环境
\usepackage{multirow}   % \multirow
\usepackage[table]{xcolor}

\AtBeginDocument{%
  }

\begin{document}

\title{Bridging Event Streams and DiT: \\Event-Guided Video Frame Interpolation}

\author{Guixu Lin}
\affiliation{%
  \institution{The University of Tokyo}
  \city{Tokyo}
  \country{Japan}}

\author{Yuyang Yu}
\affiliation{%
  \institution{South China University of Technology}
  \city{Guangzhou}
  \country{China}}

\author{Xiang Ji}
\affiliation{%
  \institution{The University of Tokyo}
  \city{Tokyo}
  \country{Japan}}

\author{Linyao Chen}
\affiliation{%
  \institution{The University of Tokyo}
  \city{Tokyo}
  \country{Japan}}

\author{Zhengwei Yin}
\affiliation{%
  \institution{The University of Tokyo}
  \city{Tokyo}
  \country{Japan}}

\author{Mengshun Hu}
\affiliation{%
  \institution{Wuhan University}
  \city{Wuhan}
  \country{China}}

\author{Mingdeng Cao}
\affiliation{%
  \institution{The University of Tokyo}
  \city{Tokyo}
  \country{Japan}}

\author{Shengfeng He}
\authornote{Corresponding authors.}
\affiliation{%
  \institution{Singapore Management University}
  \city{Singapore}
  \country{Singapore}}

\author{Yinqiang Zheng}
\authornotemark[1]
\affiliation{%
  \institution{The University of Tokyo}
  \city{Tokyo}
  \country{Japan}}

\renewcommand{\shortauthors}{Lin et al.}

\setcopyright{none}
\settopmatter{printacmref=false}
\begin{abstract}
 % Latent Diffusion Models have advanced video frame interpolation by generating intermediate frames between input frames. However, effectively handling large temporal gaps and complex motion remains a challenge, often leading to artifacts. We argue that event camera signals, with their ability to capture continuous motion at high temporal resolutions, are ideal for bridging these temporal gaps and enhancing interpolation precision. Given the impracticality of training an event-assisted model from scratch, we introduce a novel adapter-based framework that   effortlessly integrates high-temporal-resolution cues from event cameras into pre-trained image-to-video models without modifying their underlying structure. Our method leverages Image Warped Events (IWEs) and bidirectional sparse optical flow for precise spatial and temporal alignment, significantly reducing artifacts and improving interpolation quality. Experimental results demonstrate that our event-enhanced interpolation achieves superior accuracy and temporal coherence compared to existing state-of-the-art methods.
Latent diffusion models have recently advanced video frame interpolation by synthesizing intermediate frames between input images. However, handling large temporal gaps and complex motion remains challenging, often resulting in motion blur, structural distortions, and temporal inconsistencies. Event cameras provide high-temporal-resolution motion cues that are well suited for bridging these gaps and improving interpolation quality. 
To exploit this advantage without training an event-assisted model from scratch, we propose an adapter-based framework that incorporates event-derived cues into a pre-trained image-to-video diffusion model with minimal architectural changes. Specifically, our method leverages Image Warped Events (IWEs) and bidirectional sparse optical flow to provide spatially and temporally aligned guidance during generation. By injecting these event-guided structural and motion cues into the diffusion process, our approach reduces interpolation artifacts and improves both reconstruction fidelity and temporal coherence. Experimental results on real and synthetic benchmarks show that our method consistently outperforms existing state-of-the-art approaches. The project page is at \url{https://joseph-lin-tech.github.io/BridgeEventDiT-VFI/}.
 
\end{abstract}

\begin{teaserfigure}
\centering
\vspace{0mm}\begin{minipage}[t]{0.24\textwidth}
  \centering
  \includegraphics[width=\linewidth]{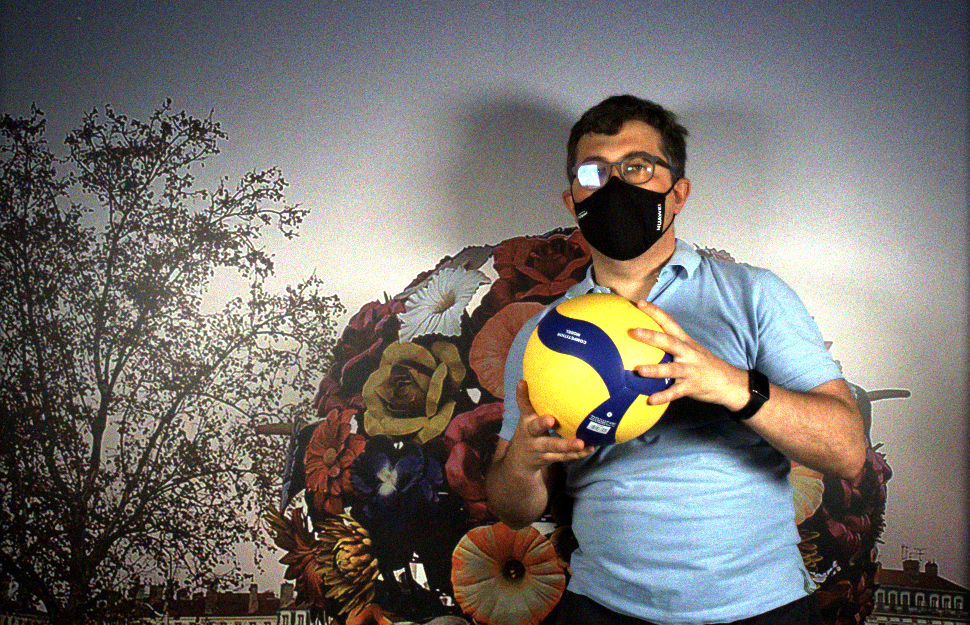}\\
 % \vspace{-1pt}
  \includegraphics[width=\linewidth]{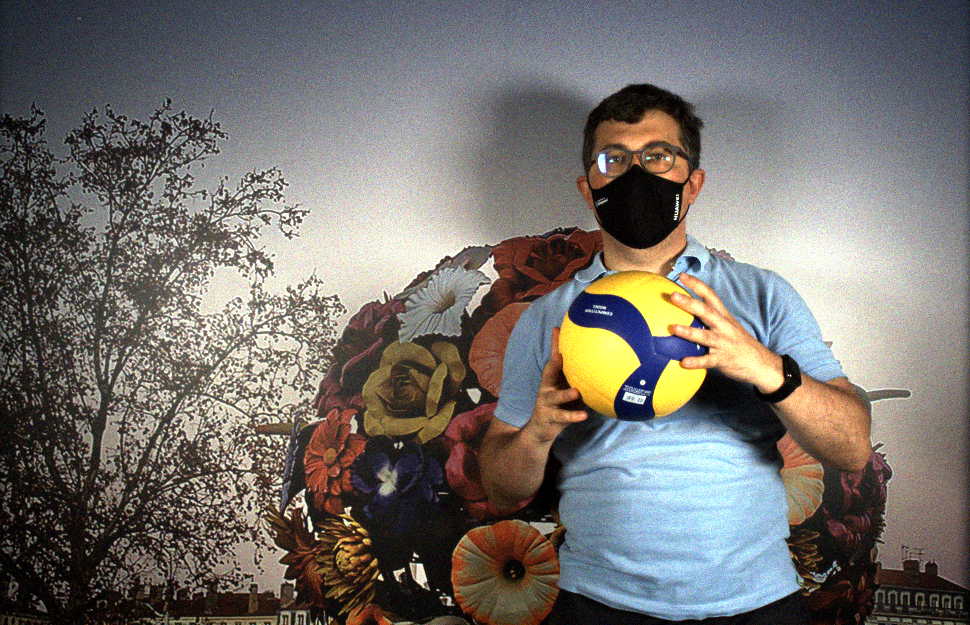}
 % \vspace{0.5mm} % 尽量不超过 -2mm
  \small {(a) Reference frames}
\end{minipage}
 %\hspace{-5pt}
\begin{minipage}[t]{0.24\textwidth}
  \centering
 \includegraphics[width=\textwidth]{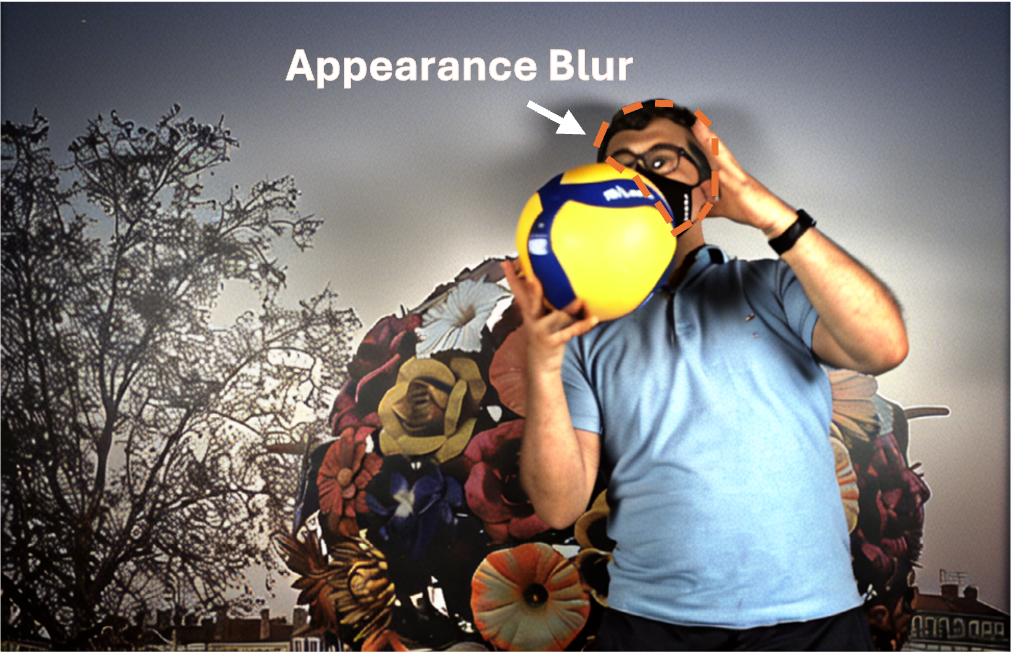}\\
  % \vspace{-1pt}
   \includegraphics[width=\textwidth]{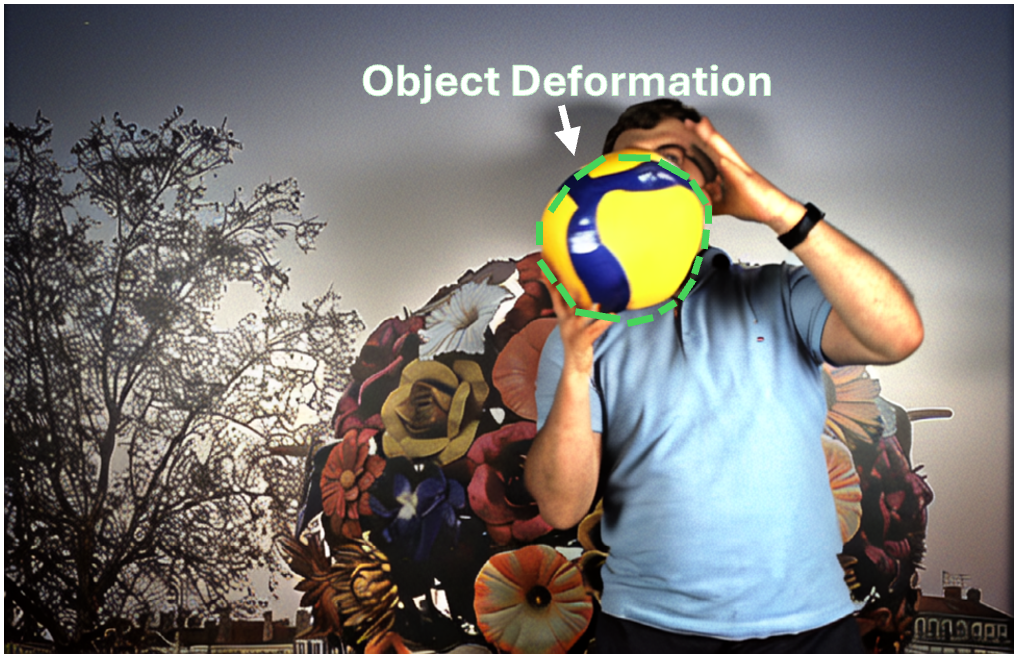}
 %\vspace{0.5mm}
\small (b) FCVG 
\end{minipage}
 %\hspace{-5pt}
\begin{minipage}[t]{0.24\textwidth}
  \centering
  \includegraphics[width=\linewidth]{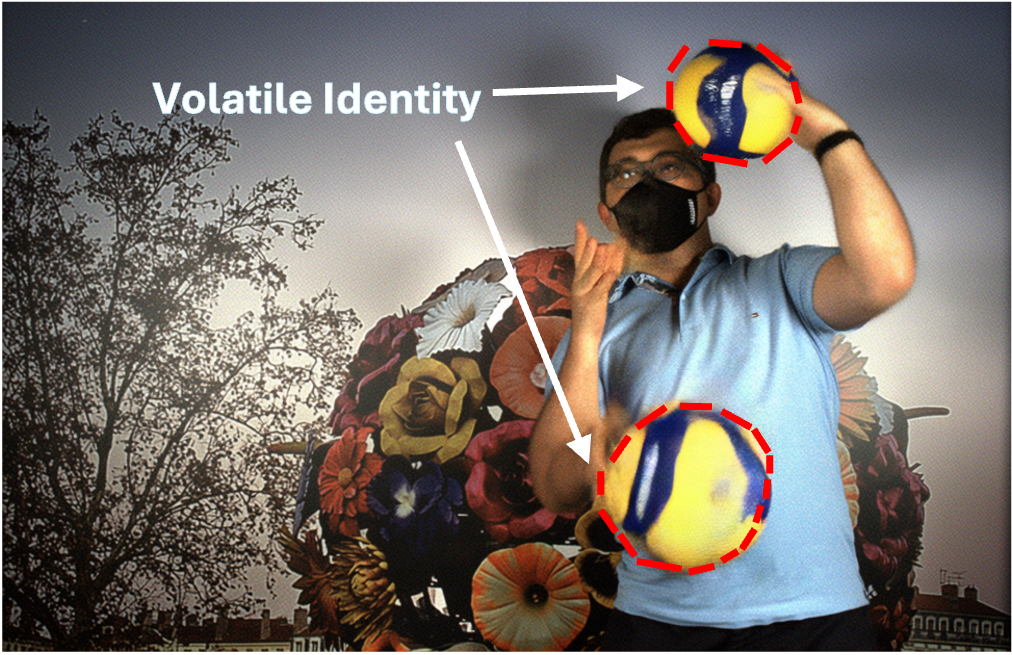}\\
 % \vspace{-1pt}
    \includegraphics[width=\linewidth]{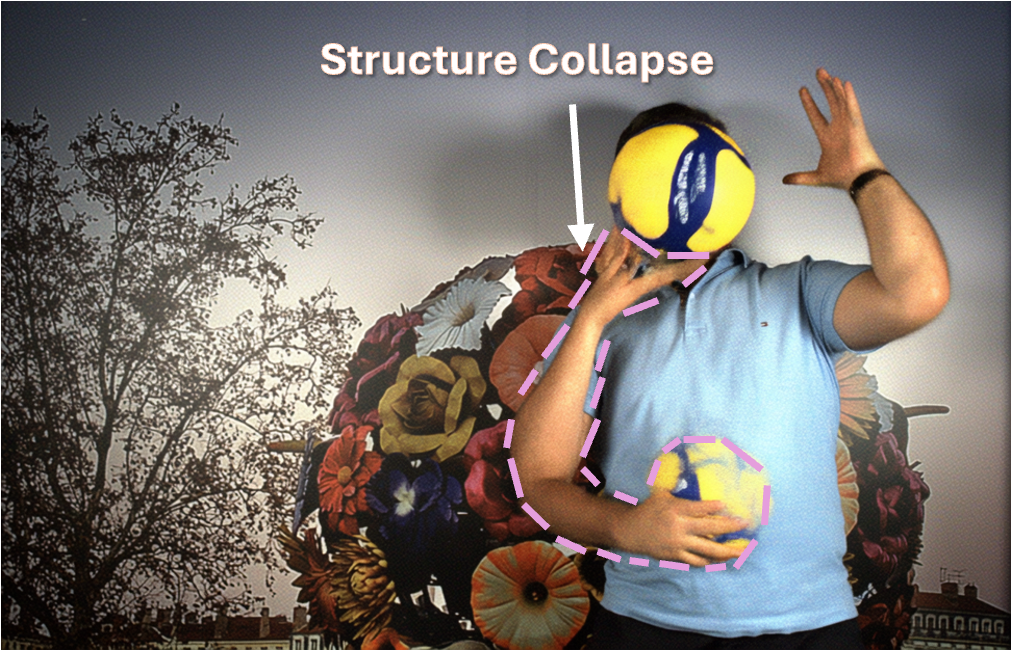}  
  %\vspace{0.5mm}
  \small (c) Wan2.1 FLF2V 
  \end{minipage}
% \hspace{-5pt}
\begin{minipage}[t]{0.24\textwidth}
  \centering
  \includegraphics[width=\linewidth]{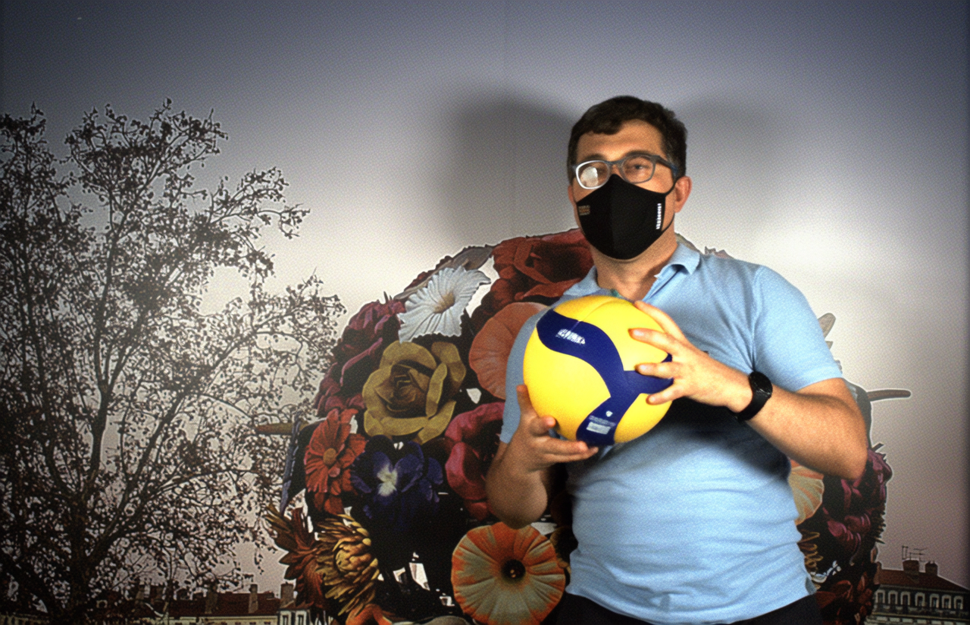}\\
  %\vspace{-1pt}
    \includegraphics[width=\linewidth]{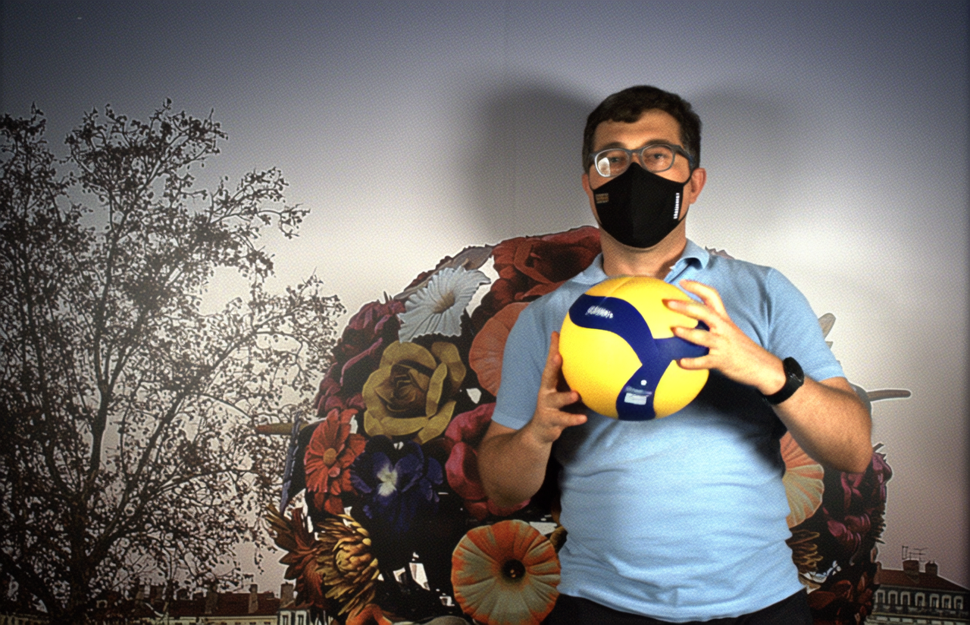}  
  %  \vspace{0.5em}
  \small (d) Ours
  \end{minipage}
  \vspace{-5pt}
    \caption{We propose an  event-augmented approach for Diffusion Transformer (DiT)-based video frame interpolation, integrating event streams to generate clear and temporally consistent frames. (a) shows the ground-truth intermediate frames at different time stamps for reference. Compared to FCVG~\cite{zhu2024generative} and Wan2.1 FLF2V~\cite{wan2025wan}, which rely solely on the start and the end frames and often result in motion blur, object deformation, identity inconsistency, and structural artifacts. In contrast, our approach produces sharper and more natural results.}
    \label{fig:teaser}
    \vspace{20pt}
\end{teaserfigure}
\maketitle
\hypersetup{
  pdfauthor={Guixu Lin, Yuyang Yu, Xiang Ji, Linyao Chen, Zhengwei Yin, Mengshun Hu, Mingdeng Cao, Shengfeng He, Yinqiang Zheng},
  pdfsubject={Event-guided video frame interpolation},
  pdfcreator={LaTeX},
  pdfpublisher={The authors}
}

\section{Introduction}
\label{sec:intro}

Latent Diffusion Models (LDMs) have recently made significant strides in both image and video generation, spurring advances in video frame interpolation, which involves synthesizing intermediate frames between start and end frames. Leveraging pre-trained image-to-video (I2V) diffusion models, recent methods can address challenges such as complex motion and large temporal gaps—scenarios where traditional interpolation techniques, which rely on motion estimation and motion compensation, often falter.

For example, recent methods such as GI~\cite{wang2024generative} and FCVG~\cite{zhu2024generative} leverage the generative capabilities of Stable Video Diffusion (SVD)~\cite{blattmann2023stable}, which is based on a U-Net diffusion architecture.  More recently, Wan2.1 FLF2V~\cite{wan2025wan} utilizes a powerful Diffusion Transformer (DiT)-based video diffusion model, achieving superior performance in interpolation across substantial temporal gaps. However, despite these advances, LDM-based interpolation methods still suffer from noticeable artifacts, particularly in the intermediate frames of the generated sequences, as illustrated in Fig.~\ref{fig:teaser}. We attribute these limitations to the reliance solely on start and end frames for guidance, constraining the interpolation quality.

Event cameras, which asynchronously capture pixel-wise brightness changes, offer unique advantages, including high temporal resolution, broad dynamic range, and low latency. Unlike traditional cameras with fixed frame rates, event cameras provide continuous, high-temporal-resolution motion information that can enrich frame interpolation, especially in complex, high-speed scenes~\cite{tulyakov2021time, tulyakov2022time, sun2023event, liu2024video}. 
%
% We posit that explicit temporal cues from event streams can significantly reduce artifacts and enable more seamless temporal transitions.
However, integrating raw event streams into the I2V diffusion models is non-trivial because event data is sparse, asynchronous, and lacks direct compatibility with the dense, grid-based representations used in mainstream generative models. Furthermore, large-scale paired event-video datasets are scarce, making end-to-end supervised training infeasible.

To address these challenges, we propose to extract Image Warped Events (IWEs) and bidirectional sparse optical flow from event streams using contrast maximization techniques~\cite{Stoffregen19cvpr, shiba2024secrets}. These representations translate the event modality into edge-like and motion-consistent cues that closely align with control signals commonly used in diffusion-based video generation, such as edge maps, flow fields~\cite{karmokar2025secrets, jiang2025vace, burgert2025go}. This serves as a conceptual and practical bridge between event-based vision and frame-based video diffusion.

Building on this insight, we propose a novel adapter-based framework that injects motion-aware signals derived from events into a pre-trained video diffusion model. Our method requires only minimal fine-tuning on limited event-video data and does not alter the underlying diffusion architecture.
Specifically, we introduce two plug-in adapters:
(1) An IWE encoder, which embeds edge-consistent spatial structure into the input latent space;
(2) A flow-based alignment-and-fusion adapter, which warps latent features using bidirectional flow before the DiT block and fuses them to form temporally aligned intermediate representations.
These adapters inject event-derived structural and temporal cues into the generative process, enhancing interpolation quality and reducing artifacts. To facilitate broader generalization and benchmarking, we also construct a large-scale synthetic event-video dataset, EvPexels, comprising 1,100 diverse scenes (about 390,000 RGB frames) spanning a wide range of motions. To the best of our knowledge, EvPexels is the largest synthetic dataset specifically designed for event-based video frame interpolation. 

In summary, this paper makes the following key contributions to event-guided video frame interpolation:
\begin{enumerate}
\item We propose an adapter-based framework that incorporates event-derived signals into DiT-based video diffusion models for frame interpolation, improving temporal consistency and reducing artifacts.
\item We bridge the gap between event streams and LDM-compatible control signals by extracting IWEs and bidirectional optical flow, enabling integration into mainstream generative pipelines.
\item We construct a synthetic event-video dataset with 1,100 diverse motion-rich scenes to support training of event-aware frame interpolation models.
\item Extensive experiments validate the effectiveness of our method, showing superior interpolation quality compared to state-of-the-art baselines.
\end{enumerate}

\section{Related Work}
This section provides an overview of research efforts closely related to our work. We begin by reviewing traditional video frame interpolation techniques, including both frame-based and event-guided approaches. We then examine recent developments in the emerging generative paradigm, with a focus on diffusion-based interpolation methods.

\label{sec:relatedwork}

\subsection{Traditional Video Frame Interpolation}
Video Frame Interpolation (VFI) is a technique used to reconstruct intermediate frames from a pair of input frames~\cite{huang2022real, kong2022ifrnet, li2023amt, zhang2023extracting, niklaus2017video, bao2019depth}. While traditional VFI methods perform well in scenarios with simple motion, they often struggle with complex motions or substantial scene changes between frames.

Event streams, which capture fine-grained motion details between frames, provide more accurate motion estimation for VFI, making event-based VFI methods increasingly popular~\cite{tulyakov2021time, tulyakov2022time, yu2021training, he2022timereplayer, zhang2022unifying, kim2023event, sun2023event, lin2023_eventguided}. For example, Time Lens~\cite{tulyakov2021time} introduced the first VFI model combining warping and synthesis-based approaches.
More recently, CBMNet~\cite{kim2023event} and TimeLens-XL~\cite{ma2024timelens} have advanced the state of the art by significantly improving the performance of event-based VFI.

While event-guided VFI has improved motion estimation accuracy, these methods still encounter challenges with significant scene changes, such as the appearance of new objects, where event data alone may be insufficient. Consequently, the performance of traditional VFI methods in real-world scenarios involving complex motion and scenes still requires further refinement, motivating diffusion model-based VFI methods, which we discuss next.

\subsection{Diffusion-based Video Frame Interpolation}
Diffusion-based VFI techniques have garnered attention due to their ability to handle large and ambiguous motions between frames more effectively than traditional methods. Early work, such as MCVD~\cite{voleti2022mcvd}, employed latent diffusion models (LDMs) for video prediction and interpolation. Building on this, LDMVFI~\cite{danier2024ldmvfi} applied LDMs specifically for frame interpolation, while VIDIM~\cite{jain2024video} advanced this by training diffusion models on larger datasets to enhance performance. CBBD~\cite{lyu2024frame} introduced the Consecutive Brownian Bridge Diffusion model, which reduces cumulative variance based on the Brownian Bridge Diffusion Model framework~\cite{li2023bbdm}. Similarly, DreamMover~\cite{shen2024dreammover} utilizes stable diffusion priors to interpolate frames with large motions.

However, most of the above diffusion-based models rely on image-to-image (I2I) diffusion frameworks, requiring complex architecture designs and specialized training on specific video datasets. This approach often overlooks recent advancements in I2V models. The emergence of large-scale I2V diffusion models offers a more efficient alternative: adapting pre-trained models (e.g., Stable Video Diffusion~\cite{blattmann2023stable}, Wan2.1~\cite{wan2025wan}) for VFI with minimal modifications, enabling training-free or tuning-free methods. Such approaches capitalize on the potential of pre-trained I2V diffusion models for video generation.
For instance, TRF~\cite{feng2024explorative} adapts a video generation model for bounded generation, using initial and final frames to synthesize intermediate frames. However, TRF~\cite{feng2024explorative} does not fully address motion consistency between frames, prompting recent work, such as GI~\cite{wang2024generative}, to introduce a reverse motion method to improve frame-to-frame coherence. 
More recently, ViBiD~\cite{yang2024vibidsampler}, FCVG~\cite{zhu2024generative} and Wan2.1 FLF2V~\cite{wan2025wan} have further advanced performance in this domain. 
Nonetheless, most prior methods have underappreciated the importance of event signals for modeling fine-grained temporal dynamics. Recent efforts, such as U-Net–based latent diffusion models~\cite{chen2024repurposing}, have begun to address this gap; however, our approach is fundamentally different. VDM-EVFI~\cite{chen2024repurposing} follows a ControlNet-style design: it converts raw event streams into event voxel grids and trains a heavy ControlNet to encode event information as conditional signals. In contrast, we transform the spatial and motion cues encoded by events into an intermediate representation that integrates with a DiT-based interpolation model, enabling effective event guidance without requiring end-to-end event-centric training.

% Once the motion adapter converges, we unfreeze the spatial self-attention layers in both the adapter and U-Net to refine color consistency. To achieve this, we introduce a Color Manifold Loss $\mathcal{L}_{CM}$, which constrains the model to learn  coherent color transitions.

\section{Methodology}
In this section, we first provide an overview of event-based video frame interpolation. We then introduce our fine-tuning pipeline, which involves extracting motion information—specifically,  IWEs and bidirectional optical flow—from raw event streams. Next, we present the design of the IWE encoder, which injects edge-aligned spatial features into the video diffusion model, and the alignment and fusion adapter, which utilizes the bidirectional optical flow to warp latent features.

\subsection{Preliminaries}

\subsubsection{I2V Models for Video Frame Interpolation}
The I2V model based on latent diffusion primarily relies on the start frame \( I_0 \) and the end frame \( I_1 \) to perform video interpolation. Let the input video sequence be denoted as  
\begin{equation}
\mathbf{I} = \{ I_0, I_{1/N}, \dots, I_{(N-1)/N}, I_1 \},
\end{equation}
where \( i \in [0, 1] \) represents normalized time steps. Following Wan2.1~\cite{wan2025wan}, we adopt a video VAE that compresses the temporal resolution by a factor of 4. Accordingly, the sequence is encoded into a latent video representation  
\begin{equation}
\mathbf{Z} = \{ z^0, z^1, \dots, z^k, \dots, z^{T-1} \},
\end{equation}
where \( T = \lfloor (N+1)/4 \rfloor + 1 \), and \( k \in [0, T-1] \).

To avoid ambiguity, all frame indices refer to the latent space throughout the remainder of this paper, unless otherwise specified.
The base video diffusion model we use, FLF2V in Wan2.1~\cite{wan2025wan}, is trained to predict a constant velocity vector \( v_t \) from a noisy latent representation \( x_t^k \), the timestamp \( t \), and the corresponding text condition \( c_{\text{txt}} \). The training objective is formulated as the mean squared error (MSE) between the predicted velocity \( u(x_t^k, c_{\text{txt}}, t; \theta) \) and the ground-truth \( v_t \):
\begin{equation}
\mathcal{L} = \mathbb{E}_{k, t, c_{\text{txt}}} \left\| u(x_t^k, c_{\text{txt}}, t; \theta) - v_t \right\|^2,
\end{equation}
where \( \theta \) denotes the model parameters. This objective guides the model to learn continuous trajectories in latent space, conditioned on the input prompt.

\subsubsection{Event-assisted Video Frame Interpolation}
To enhance video frame interpolation, we incorporate an event stream  
\begin{equation}
    \mathcal{E} = \{ e_i = (x_i, y_i, \tau_i, p_i) \}
\end{equation}
between frames \( I_0 \) and \( I_1 \). Each event \( e_i \in \mathcal{E} \) occurs at spatial coordinates \( (x_i, y_i) \), at time \( \tau_i \), with polarity \( p_i \in \{-1, +1\} \), capturing sparse spatiotemporal changes in the scene.
Accordingly, the training objective for the event-assisted interpolation task is modified to:
\begin{equation}
\mathcal{L} = \mathbb{E}_{k, t, \mathcal{E}, c_{\text{txt}}} \left\| u(x_t^k, \mathcal{E}, c_{\text{txt}}, t; \theta) - v_t \right\|^2.
\end{equation}
We first apply contrast maximization (CMax) to compute bidirectional optical flows \( \mathbf{f}_{k-1 \rightarrow k} \) and \( \mathbf{f}_{k+1 \rightarrow k} \) from the event intervals \([k{-}1, k]\) and \([k{+}1, k]\), respectively:
\begin{equation}
\begin{aligned}
\mathbf{f}_{k-1 \rightarrow k},\; \mathcal{W}^{k-1 \rightarrow k} &= \text{CMax}(\mathcal{E}_{[k-1, k]}), \\
\mathbf{f}_{k+1 \rightarrow k},\; \mathcal{W}^{k+1 \rightarrow k} &= \text{CMax}(\mathcal{E}_{[k+1, k]}),
\end{aligned}
\end{equation}
where \( \mathcal{E}_{[k-1, k]} \) and \( \mathcal{E}_{[k+1, k]} \) denote the subsets of events occurring between the respective frames. The outputs \( \mathcal{W}^{k-1 \rightarrow k} \) and \( \mathcal{W}^{k+1 \rightarrow k} \) are IWE representations warped by forward and backward optical flow.
We employ an IWE encoder to extract the aligned spatial edge features from the IWE maps, denoted as \( \mathbf{F}_{\mathcal{W}}^k \), which are subsequently injected into the input latent representation to enhance structural guidance.

In the alignment and fusion module, we warp the DiT features from the adjacent latent states \( \mathbf{F}_{x_t}^{k-1} \) and \( \mathbf{F}_{x_t}^{k+1} \) toward the current frame \( k \), using the estimated optical flows:
\begin{equation}
\begin{aligned}
\mathbf{F}_{x_t}^{k-1 \rightarrow k} &= \text{Warp}(\mathbf{F}_{x_t}^{k-1},\; \mathbf{f}_{k \rightarrow k-1}), \\
\mathbf{F}_{x_t}^{k+1 \rightarrow k} &= \text{Warp}(\mathbf{F}_{x_t}^{k+1},\; \mathbf{f}_{k \rightarrow k+1}).
\end{aligned}
\end{equation}
These aligned features are then fused with the center frame's features \( \mathbf{F}_{x_t}^{k} \) via a fusion function \( G(\cdot) \):
\begin{equation}
\mathbf{F}^{k}_{\text{fused}} = G\left( \mathbf{F}_{x_t}^{k},\; \mathbf{F}_{x_t}^{k-1 \rightarrow k},\; \mathbf{F}_{x_t}^{k+1 \rightarrow k} \right).
\end{equation}
Finally, the fused residual features of all frames \( \mathbf{F}_{\text{fused}} \) are added to the DiT features to enhance temporal consistency across frames.

\subsection{Framework}
\begin{figure*}[t]
  \centering
\includegraphics[width=1\linewidth]{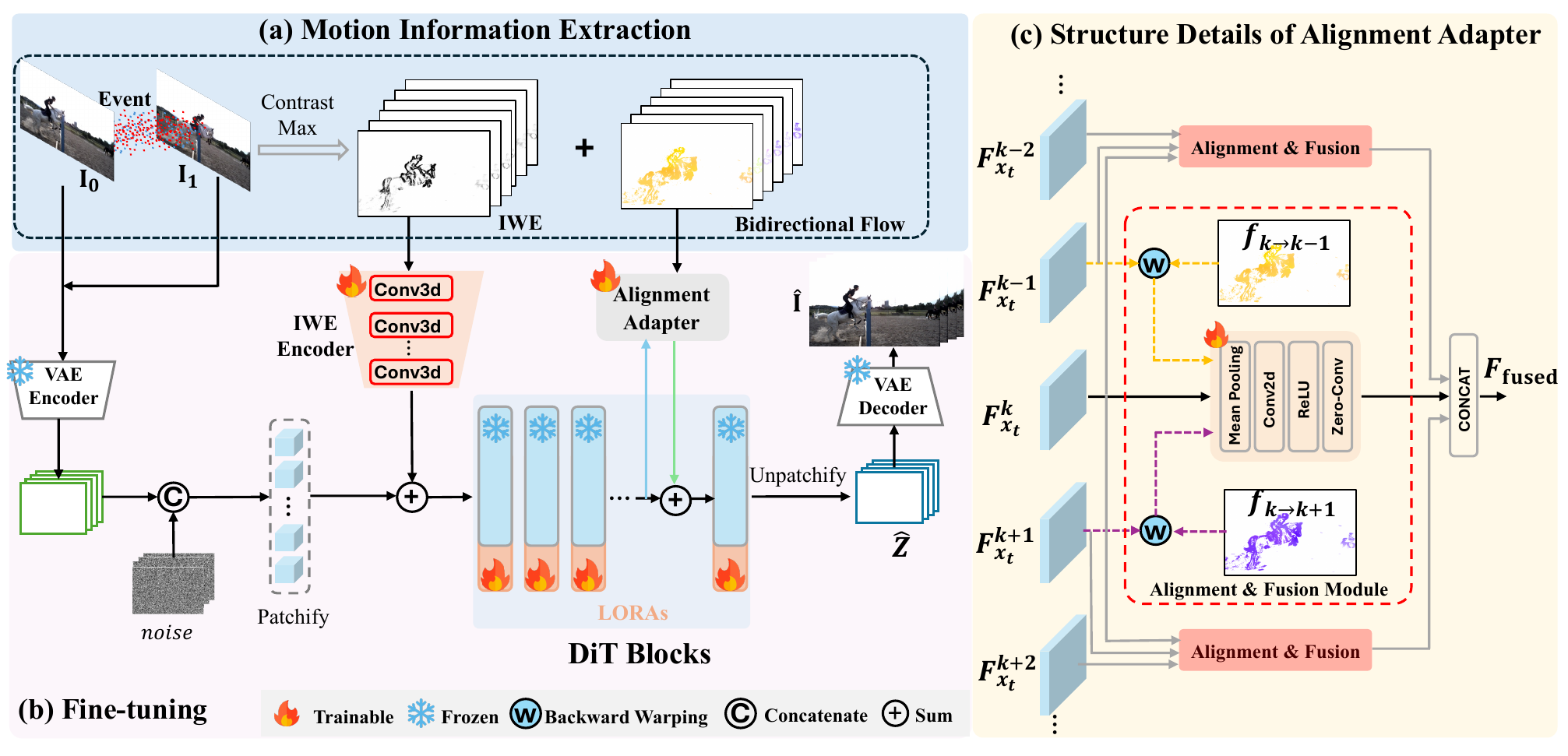}
\vspace{-10pt}
 \caption{Illustration of Our Framework. (a) We extract bidirectional sparse optical flow and  IWEs from the input event stream using the Contrast Maximization (CMax) method.
(b) During fine-tuning, the model is enhanced with three components: an IWE encoder, alignment adapters inserted into a subset of DiT blocks, and LoRA layers applied to all DiT blocks.
(c) The flow-based alignment adapter leverages the bidirectional flows to warp intermediate features from neighboring frames, aligning them temporally with the current frame. This facilitates motion-consistent feature propagation throughout the denoising process.}
  \label{fig:framework}
  \vspace{10pt}
  \end{figure*}

As shown in Fig.~\ref{fig:framework}, our framework takes the start frame ${I}_{0}$, the end frame ${I}_{1}$, and the corresponding event stream as input, and outputs the intermediate frames $\hat{\mathbf{I}}$. The overall architecture consists of an event representation module and an adapter-enhanced fine-tuning strategy.

\subsubsection{Event Representation}
To extract motion cues from the event stream, we follow the principle of contrast maximization~\cite{Stoffregen19cvpr, shiba2024secrets}, which enables both optical flow estimation and generation of the Image Warped Events (IWE)—a sharp, edge-aware image obtained by temporally aligning events at a designated reference time. We employ an off-the-shelf CMax-based method~\cite{shiba2024secrets} to compute optical flows and their corresponding IWEs.
For temporal alignment, we divide the event stream between \( I_0 \) and \( I_1 \) into multiple temporal segments. For each segment \([k{-}1, k]\), we compute a sparse forward optical flow \( \mathbf{f}_{k-1 \rightarrow k} \) and the corresponding IWE \( \mathcal{W}^{k-1 \rightarrow k} \), by masking the dense flow with event activity. Similarly, we reverse the event stream and extract backward flows \( \mathbf{f}_{k+1 \rightarrow k} \) and IWEs \( \mathcal{W}^{k+1 \rightarrow k} \). This process yields bidirectional sparse optical flows and edge-aware IWE representations across the entire sequence.

\subsubsection{IWE-Based Spatial Conditioning}
% We design an encoder to extract aligned structured features from the bi-directional IWEs, resulting in the spatial aligned IWE features $\mathbf{F}_\mathcal{W}^{k}$, where k$in$[0,T].
% To make the DiT blocks can fit with the IWE features, we use the LORA manner to make the video diffusion model fit the IWE inputs.
To extract structured spatial cues from the event stream, we design an encoder that processes the bidirectional IWEs and produces aligned feature maps \( \mathbf{F}_\mathcal{W}^{k} \), where \( k \in [0, T] \). IWEs are known to correlate strongly with scene edges and object boundaries~\cite{karmokar2025secrets}, making them particularly effective for guiding frame synthesis in motion-intensive regions.
Given the sparse and lightweight nature of the IWE signal, we adopt a simple yet effective integration strategy. The IWE maps are encoded into feature representations using a compact network composed of 3D convolutional layers. These features are then injected into the video latent space via element-wise addition to the input latents of the diffusion model. This design enables edge-aware spatial conditioning with minimal computational overhead.
To adapt the pre-trained diffusion model to this new input modality, we apply LoRA-based fine-tuning across all DiT blocks. This allows the model to leverage IWE-derived structural features while keeping the majority of the original parameters frozen, ensuring parameter efficiency and architectural compatibility.

% \subsubsection{Event-assisted Adapter For Motion Guidance}

% \subsubsection{Bidirectional Optical Flow Warping $\&$ Fusion Network:}
% After the DiT block, we rearrange the patches into the frames, which has the spatial corresponding with the optical flows.
% We utilize bidirectional optical flows, specifically $f_{{k} \rightarrow {k-1}}$ and $f_{{k} \rightarrow {k+1}}$ to warping the previous and next frame feature to the current frame, to align the temporal features, These three features are processed mean pooling and the input the fusion network, which consists of  convolutional layers. The resulting features are then added into DiT features. And after the summation, we rearrange the features into patches to feed to the next DiT block.
 
\subsubsection{Flow-Based Temporal Alignment and Fusion}  
Before each selected DiT block, we rearrange the patch-based latent representations back into a frame-wise format that is spatially aligned with the optical flows. To enforce temporal consistency, we first reshape the bidirectional optical flows and warp the features from the previous and next frames toward the current frame using the corresponding forward and backward flows.
The three aligned features (from frames \( k{-}1 \), \( k \), and \( k{+}1 \))  are then aggregated via mean pooling and passed through a lightweight fusion network composed of convolutional layers.   The fused feature is subsequently added to the original DiT feature as a residual correction, enhancing both spatial fidelity and temporal coherence.
Finally, the updated frame-wise features are rearranged back into patch tokens and propagated to the selected DiT block for further refinement. This design allows the model to benefit from explicit motion guidance without incurring significant computational overhead. To balance performance and efficiency, the alignment-and-fusion module is only applied to a subset of DiT blocks rather than all layers.

\section{Experiments}
 
\subsection{Experimental Setup}

\subsubsection{Training Datasets}
We fine-tune our framework using two datasets: the real-world BS-ERGB dataset~\cite{tulyakov2022time} and our curated synthetic EvPexels dataset.
The BS-ERGB dataset contains high-speed image-event pairs captured at a resolution of 970×625 and a frame rate of 28 fps. The training split includes 48 video clips, while the test split consists of 26 clips and is used for quantitative evaluation.

To augment training diversity, we introduce the EvPexels dataset, constructed from videos collected via the Pexels platform ({{$https://www.pexels.com$}}). We select videos exhibiting diverse motion patterns using TransNet V2~\cite{soucek2024transnet}, ensuring single-shot segments suitable for the frame interpolation task. Event streams are synthesized from RGB videos using the Vid2e simulator~\cite{gehrig2020video}. The resulting dataset comprises 1,100 video sequences, totaling 389,761 frames at a resolution of 704×480. 
The visualization of the EvPexels dataset are provided in the supplementary materials.

\subsubsection{Test Datasets}
We evaluate performance on a real-captured dataset (BS-ERGB test set) and two synthetic datasets.
As for the two additional synthetic datasets, following prior works such as TRF~\cite{feng2024explorative} and GI~\cite{wang2024generative}, we select 50 video clips from the DAVIS dataset~\cite{pont20172017} and 30 clips from Pexels, each consisting of 25 frames. These datasets cover diverse motion scenarios and provide a comprehensive evaluation of interpolation quality.

\subsubsection{Implementation Detail}
We adopt the open-source FLF2V model from Wan2.1~\cite{wan2025wan} as our base video diffusion architecture. The model takes the first and last frames as input and generates a video of fixed length (81 frames) in latent space. The learning rate is set to \( 1 \times 10^{-4} \), and all input images are resized to a resolution of 832$\times$480 during training. Other hyperparameters follow the original FLF2V configuration without modification. In our experiments, optical flow information is injected into two DiT blocks to enhance motion guidance.
 We fine-tune our model for 4,000 steps on 8 NVIDIA A800 GPUs with a global batch size of 8.

\subsubsection{Evaluation Metrics}
We calculate metrics including  PSNR (Peak Signal-to-Noise Ratio), SSIM (Structural Similarity Index Measure), Learned Perceptual Image Patch Similarity (LPIPS), Fréchet Inception Distance (FID), and Fréchet Video Distance (FVD).

\subsection{Quantitative \& Qualitative Evaluation}
\subsubsection{Quantitative Evaluation}
% To comprehensively evaluate our method, we compare it against a range of video frame interpolation (VFI) baselines. For event-based VFI, we include TimeLens~\cite{tulyakov2021time}, CBMNet-Large~\cite{kim2023event}, and the recently proposed TimeLens-XL~\cite{ma2024timelens} and VDM-EVFI~\cite{chen2024repurposing}. For frame-based VFI, we consider several diffusion-based generative approaches: TRF~\cite{feng2024explorative}, GI~\cite{wang2024generative}, ViBiD~\cite{yang2024vibidsampler},  FCVG~\cite{zhu2024generative} and Wan2.1 FLF2V~\cite{wan2025wan}.
% %
% For methods with official training code, including Wan2.1 FLF2V~\cite{wan2025wan}, VDM-EVFI~\cite{chen2024repurposing}, CBMNet-Large~\cite{kim2023event} and TimeLens-XL~\cite{ma2024timelens}, we fine-tuning the official pretrained models on our training datasets for fair comparison. For methods without publicly available training code, such as TimeLens~\cite{tulyakov2021time}, GI~\cite{wang2024generative}, ViBiD~\cite{yang2024vibidsampler}, and FCVG~\cite{zhu2024generative}, we rely on their official checkpoints for inference.

To comprehensively evaluate our method, we compare it against a broad range of video frame interpolation (VFI) baselines. For event-based VFI, we include TimeLens~\cite{tulyakov2021time}, CBMNet-Large~\cite{kim2023event}, as well as the more recent TimeLens-XL~\cite{ma2024timelens} and VDM-EVFI~\cite{chen2024repurposing}. For frame-based VFI, we consider 
restoration-based method, RIFE~\cite{huang2022real}, and 
several diffusion-based generative approaches, including TRF~\cite{feng2024explorative}, GI~\cite{wang2024generative}, ViBiD~\cite{yang2024vibidsampler}, FCVG~\cite{zhu2024generative}, and Wan2.1 FLF2V~\cite{wan2025wan}.
For methods with publicly available training code—namely Wan2.1 FLF2V~\cite{wan2025wan}, CBMNet-Large~\cite{kim2023event}, and TimeLens-XL~\cite{ma2024timelens}—we fine-tune the official pretrained models on our training datasets to ensure a fair comparison. 
For methods without released training code, such as TimeLens~\cite{tulyakov2021time}, TRF~\cite{feng2024explorative}, GI~\cite{wang2024generative}, ViBiD~\cite{yang2024vibidsampler}, and FCVG~\cite{zhu2024generative}, we use their official checkpoints for inference.
Specifically, the original VDM-EVFI~\cite{chen2024repurposing} is built upon the SVD~\cite{blattmann2023stable} backbone. For a fair comparison, we adapt VDM-EVFI to the Wan2.1 FLF2V~\cite{wan2025wan} backbone and refer to this variant as VDM-EVFI-Wan2.1. We train this model from scratch on our training set, strictly following the same training configuration as our method, using 8 NVIDIA GPUs and the same number of training iterations.
Following standard practice in diffusion-based interpolation~\cite{feng2024explorative,wang2024generative}, all methods are evaluated under a $\times 24$ interpolation setting across three datasets: BS-ERGB, DAVIS, and Pexels. The quantitative results on the BS-ERGB test set with $\times 24$ interpolation are summarized in Tab.~\ref{table:comparison_result_on_bsergb}, while the results on DAVIS and Pexels are reported in Tab.~\ref{table:comparison_result_on_syntheticdata}. 

In the $\times 24$ interpolation setting, our method achieves the best overall performance across all metrics—PSNR, SSIM, LPIPS, FID, and FVD—on the DAVIS and Pexels datasets. On BS-ERGB test dataset, our approach achieves state-of-the-art results on perceptual metrics (LPIPS, FID, and FVD), while ranking third and second in PSNR and SSIM, respectively, among distortion-based metrics. Traditional event-based methods, such as TimeLens~\cite{tulyakov2021time} and CBMNet-Large~\cite{kim2023event}, achieve higher PSNR scores, with CBMNet-Large also showing strong SSIM performance on BS-ERGB. This is largely due to the design of conventional interpolation methods, which emphasize pixel-level reconstruction accuracy. Consequently, they perform well on distortion-based metrics (e.g., PSNR, SSIM), but often fall short in generating perceptually realistic or temporally consistent frames. By contrast, generative models prioritize visual realism and temporal consistency, leading to superior perceptual quality even if pixel-wise similarity is sometimes compromised. The qualitative comparisons further illustrate these trends.

\begin{table}[ht]
\small
\centering
\setlength{\tabcolsep}{1pt}
\renewcommand{\arraystretch}{1.2}
\setlength{\abovecaptionskip}{2pt} 
\caption{Quantitative comparison of the VFI performance on the BS-ERGB   test dataset. \textbf{Bold} indicates the best performance under the $\times$24 interpolation setting.}
\begin{center}
\begin{tabular}{l|ccccc}
\hline
\rowcolor{gray!10}&
   \multicolumn{5}{c}{\textbf{BS-ERGB}}\\  \cline{2-6}
\rowcolor{gray!10} 
\textbf{Methods} &  {PSNR$\uparrow$} & {SSIM$\uparrow$} & LPIPS$\downarrow$& FID$\downarrow$ & FVD$\downarrow$\\ \hline
RIFE~\cite{huang2022real}&   22.174& 0.641& 0.172  & 35.347  & 1113.496  \\
 TRF~\cite{feng2024explorative}& 14.078 & 0.4117 &   0.426  & 47.146   &  971.424   \\ 
GI~\cite{wang2024generative}& 16.964 & 0.518& 0.311 &  33.082  & 588.371  \\ 
ViBiD~\cite{yang2024vibidsampler}&  15.525& 0.475  & 0.352  &  39.027  & 788.652 \\
FCVG~\cite{zhu2024generative}&  17.809&  0.546 & 0.302   & 26.832&726.752   \\
Wan2.1-FLF2V~\cite{wan2025wan}& 18.698 & 0.618& 0.212&18.607 & 376.828  \\
TimeLens~\cite{tulyakov2021time}&   {24.704} & {0.699} & {0.165} & 43.808   & 851.523  \\
CBMNet-Large~\cite{kim2023event}&   \textbf{25.306}& \textbf{0.712}   & 0.169  & 17.658   & 228.753    \\  
TimeLens-XL~\cite{ma2024timelens} &  21.737&  0.678 &   0.248 & 47.155   & 710.688  \\ 
VDM-EVFI-Wan2.1~\cite{chen2024repurposing} &22.402 & 0.673 & 0.282 & 15.693& 145.067\\
{Ours}  &  23.261  &  0.704 & \textbf{0.132} & \textbf{8.168}   & \textbf{117.368}  \\
\hline
\end{tabular}
\end{center}
\label{table:comparison_result_on_bsergb}
\end{table} %
% In the $\times 12$ interpolation setting, our method consistently outperforms VDM-EVFI~\cite{chen2024repurposing} across nearly all evaluation metrics, particularly those based on perception. While our approach achieves a slightly lower PSNR (0.16 dB less) on BS-ERGB, it demonstrates significantly stronger generalization ability, as evidenced by superior results on the DAVIS and Pexels datasets. Additional visual comparisons in the Appendix~\ref{visual_comparison_vdm-evfi}  further validate the effectiveness of our method.

\begin{figure*}[!ht] %{l}{0.44\linewidth}
    \centering
    \begin{minipage}{0.01\linewidth}
    \vspace{25pt}
    \centerline{\small {\rotatebox{90}{GT}}}
    \vspace{+35pt}
    \centerline{\small {\rotatebox{90}{TimeLens}}}
    \vspace{+25pt}
    \centerline{\small {\rotatebox{90}{CBMNet-Large}}}
     \vspace{+28pt}
    \centerline{\small {\rotatebox{90}{\textbf{Ours}}}}
     \vspace{25pt}
     \end{minipage}
    \hspace{-4pt}
    \begin{minipage}{0.92\linewidth}
    \centerline{ \includegraphics[width=\linewidth]{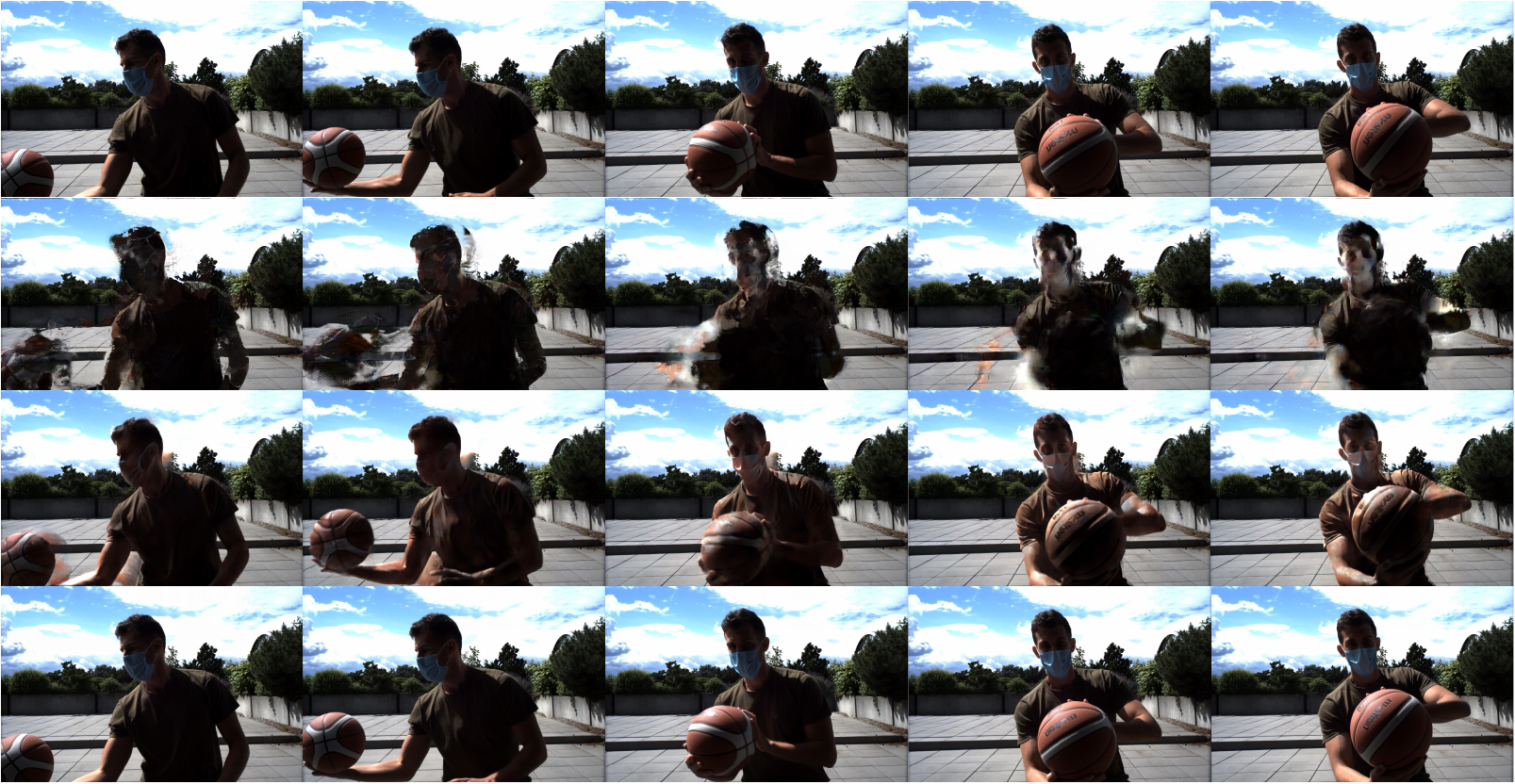}}   
    \vspace{-4pt}
    \centerline{\small  $i=3$\qquad\qquad\qquad\qquad\qquad$i=6$ \qquad\qquad\qquad\qquad\qquad $i=12$ \qquad\qquad\qquad \qquad\qquad$i=16$ \qquad\qquad\qquad\qquad$i=18$   }
    \end{minipage}
    \vspace{-3pt}
    \caption{Visual comparison of  VFI methods on the BS-ERGB test dataset (time $\times$ 24).}
    \label{fig:vis_on_bsregb}
    \vspace{0pt}
\end{figure*} 

\begin{table*}[h!]
\small
\centering
\setlength{\tabcolsep}{1pt}
\renewcommand{\arraystretch}{1.2}
\setlength{\abovecaptionskip}{3pt} 
\caption{Quantitative comparison on the VFI tasks on DAVIS and Pexels datasets (time $\times$ 24).}
\begin{center}
\begin{tabular}{l|ccccc| ccccc}
\hline
\rowcolor{gray!10}
 &\multicolumn{5}{c|}{\textbf{DAVIS}} &\multicolumn{5}{c}{\textbf{Pexels}} \\ \cline{2-11}
\rowcolor{gray!10}
\textbf{Methods} & {PSNR$\uparrow$} & {SSIM$\uparrow$} & LPIPS$\downarrow$ &    FID$\downarrow$ & FVD$\downarrow$  &      {PSNR$\uparrow$} & {SSIM$\uparrow$} &LPIPS$\downarrow$& FID$\downarrow$ & FVD$\downarrow$  \\ \hline
%RIFE~\cite{huang2022real} & 18.287  & 0.487  & 0.402   & 82.137   & 2028.480 & 21.820 & 0.634 & 0.274 & 80.118& 1976.327 \\ 
RIFE~\cite{huang2022real} & 18.287  & 0.487  & 0.402   & 82.137   & 2028.480 & 21.820 & 0.634 & 0.274 & 80.118& 1976.327 \\ 
 TRF~\cite{feng2024explorative} & 14.132& 0.459 & 0.484  & 70.528   & 1373.954   &  16.737 & 0.600 & 0.400 & 109.516 & 1624.791  \\ 
 GI~\cite{wang2024generative} & 14.850 & 0.467 & 0.406   & 55.067   & 1158.330  &   17.700 & 0.600 & 0.306 & 109.029 & 1212.097\\ 
ViBiD~\cite{yang2024vibidsampler} & 14.811 & 0.456  & 0.448  & 55.343   &1194.670  &  17.413& 0.588&0.365 & 104.089 & 1335.211\\
FCVG~\cite{zhu2024generative} &16.162 &  0.509 & 0.385  & 48.839   & 1246.823   &  19.172&0.635&0.275&105.617&1481.806 \\
Wan2.1-FLF2V~\cite{wan2025wan} &17.510 & 0.538& 0.310 &36.740 & 800.613 &19.747& 0.642&0.223&46.009 & 959.256 \\
%DynamicAnimate
TimeLens~\cite{tulyakov2021time} & {22.913} & 0.632 & 0.352 & 102.191   & 1706.523    & 27.071 & 0.757 & 0.215 & 79.886 & 1093.200\\ 
CBMNet-Large~\cite{kim2023event} & 20.633  & 0.742   &  0.343  &   79.459 & 1164.145 & 23.429  & 0.799 & 0.292 & 81.449&840.820  \\ 
TimeLens-XL~\cite{ma2024timelens}  & 17.498 & 0.530  & 0.235  & 100.506   & 1438.467   & 26.241 & 0.789 & 0.224&82.760 & 557.176\\ 
VDM-EVFI-Wan2.1~\cite{chen2024repurposing} &25.040 & 0.779 & 0.123& 15.361 & 165.236 & 29.042 & 0.821&0.082& 17.275& 158.183\\
{Ours}  &   \textbf{25.544 } & \textbf{0.799 } & \textbf{0.115 }  & \textbf{13.367 }  & \textbf{158.557}  & \textbf{ 29.089} & \textbf{0.858 } & \textbf{0.080 } & \textbf{16.319 } & \textbf{151.345}\\
\hline
\end{tabular}
\end{center}
\vspace{-5pt}
\label{table:comparison_result_on_syntheticdata}
\end{table*}

\subsubsection{Qualitative Results}
\begin{figure*}[ht!]
\centering
\begin{minipage}{0.01\linewidth}
\vspace{18pt}
\centerline{\small {\rotatebox{90}{VDM-EVFI-Wan2.1}}}
 \vspace{+60pt}
\centerline{\small {\rotatebox{90}{Ours}}}
 \vspace{+65pt}
\centerline{\small {\rotatebox{90}{{GT}}}}
 \vspace{60pt}
 \end{minipage}
\hspace{-4pt}
\begin{minipage}{0.98\linewidth}
\centerline{ \includegraphics[width=\linewidth]{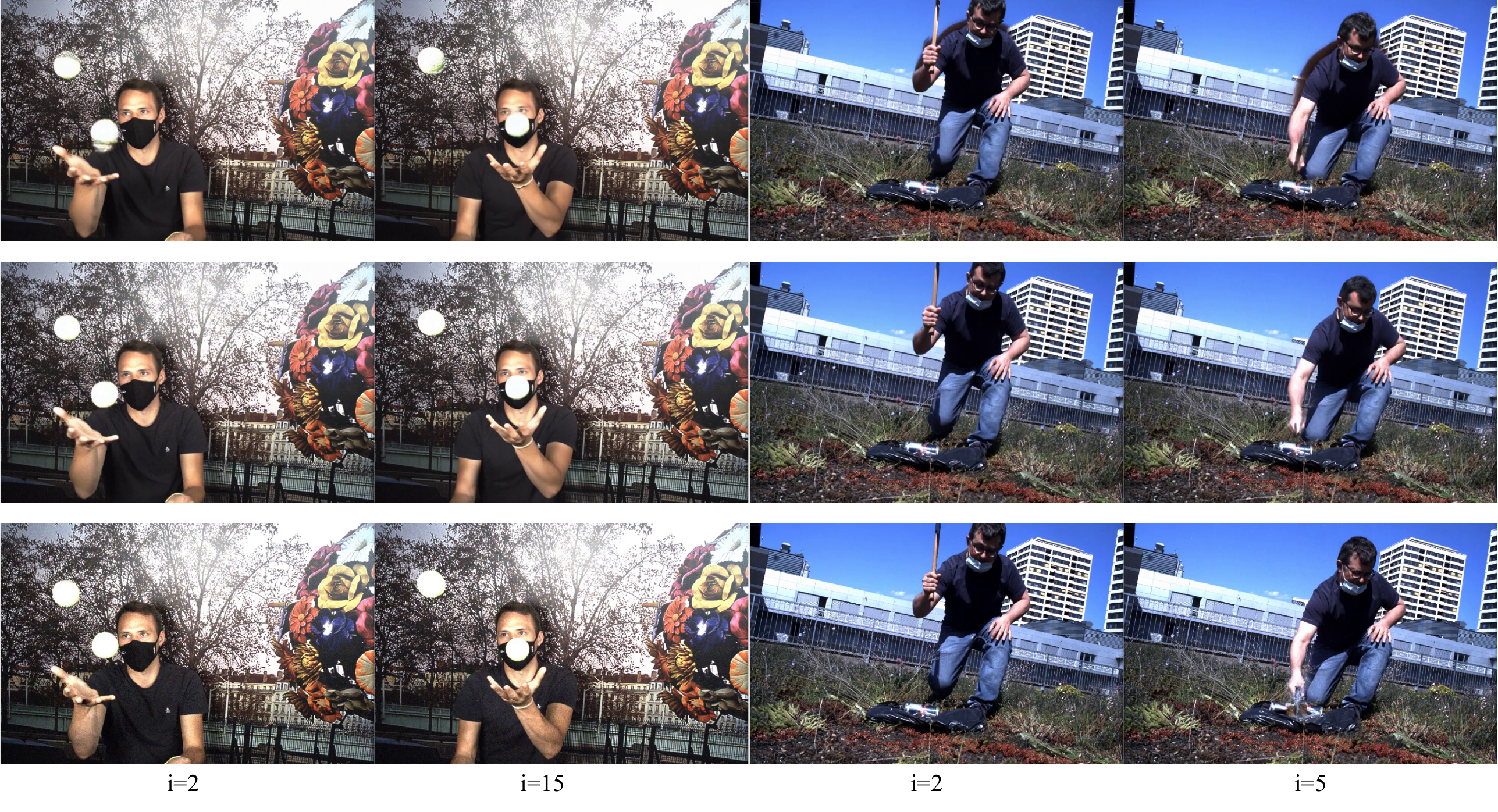}}   
\vspace{-2pt}
\end{minipage}
\vspace{-25pt}
\caption{Visual comparison between VDM-EVFI-Wan2.1 and our method on the BS-ERGB dataset.(time $\times$ 24).}
\label{fig:vis1_on_rebuttal}
%\vspace{}
\end{figure*}

\begin{figure*}[ht!]
    \centering
    \begin{minipage}{0.01\linewidth}
    \vspace{8pt}
    \centerline{ {\rotatebox{90}{VDM-EVFI-Wan2.1}}}
     \vspace{+50pt}
    \centerline{ {\rotatebox{90}{Ours}}}
     \vspace{+70pt}
    \centerline{ {\rotatebox{90}{{GT}}}}
     \vspace{60pt}
     \end{minipage}
    \hspace{-2pt}
    \begin{minipage}{0.97\linewidth}
    \centerline{ \includegraphics[width=\linewidth]{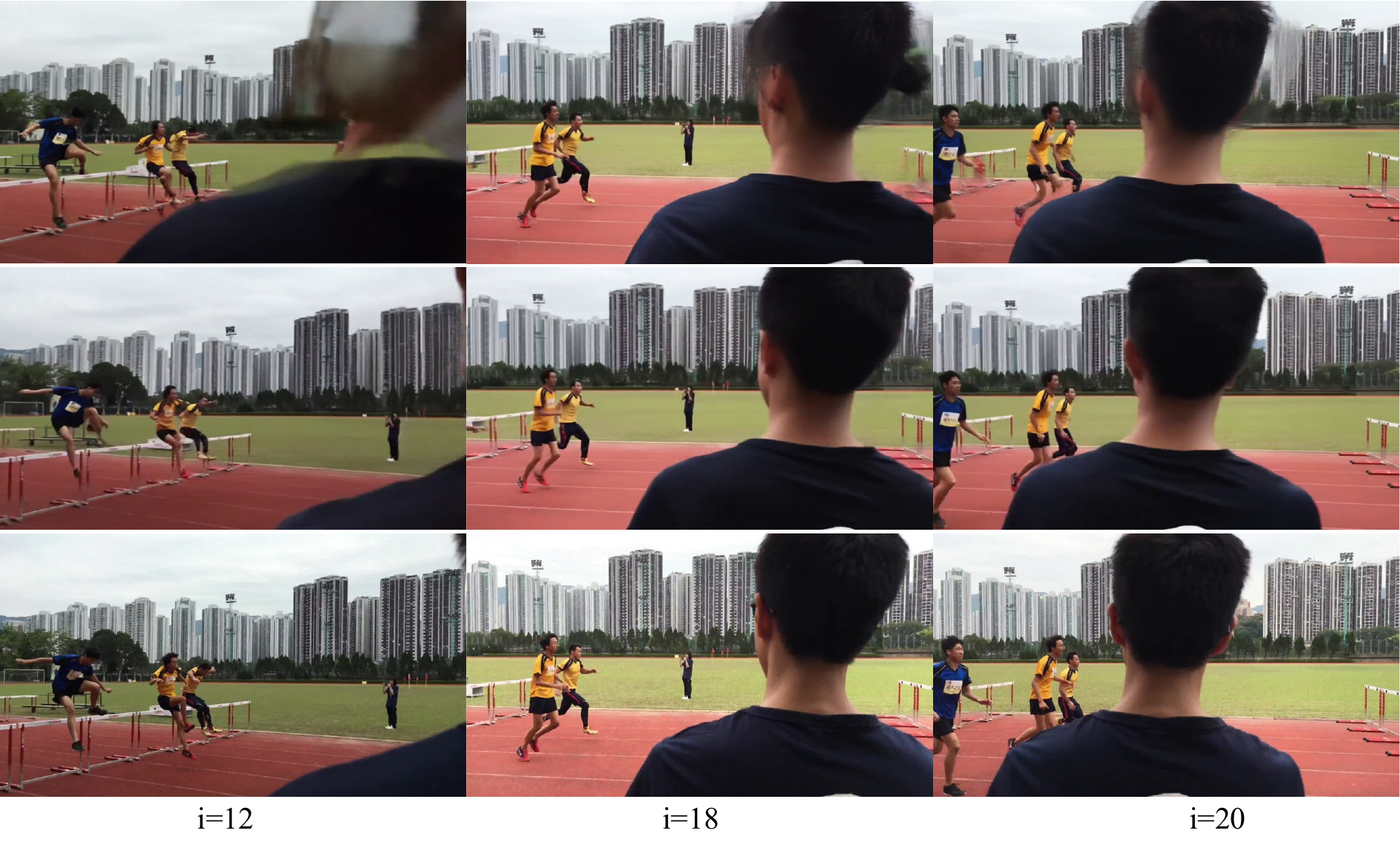}}   
    \vspace{-2pt}
    % \centerline{\small  $i=5$\qquad\qquad\qquad\quad\qquad$i=7$ \qquad\qquad\qquad\qquad\;$i=8$\qquad\qquad\qquad\qquad\;$i=9$ }
    \end{minipage}
    \vspace{-12pt}
    \caption{Visual comparison between VDM-EVFI-Wan2.1 and Ours method on the Davis dataset (time $\times$ 24).}
    \vspace{-15pt}
    \label{fig:vis2_on_rebuttal}
    %\vspace{}
    \end{figure*}

Fig.~\ref{fig:vis_on_bsregb} shows qualitative comparisons on the BS-ERGB test set, which features challenging motion involving a person and a basketball. Although TimeLens~\cite{tulyakov2021time} and CBMNet-Large~\cite{kim2023event} attain higher PSNR scores, their visual quality is clearly inferior. TimeLens~\cite{tulyakov2021time} suffers from noticeable artifacts near moving objects, while CBMNet-Large~\cite{kim2023event} generates a distorted appearance of the basketball.
In contrast, our method effectively leverages intermediate event information to accurately model the motion of dynamic foreground objects, resulting in temporally coherent and visually faithful reconstructions.
Qualitative visual comparisons with VDM-EVFI-Wan2.1 are shown in Figs.~\ref{fig:vis1_on_rebuttal} and~\ref{fig:vis2_on_rebuttal}, demonstrating the superior visual quality of our method and confirming the effectiveness of our event representation and injection strategy.

%\begin{wraptable}{r}{0.5\textwidth}%
% \begin{table}[ht!]
% \small
% \centering
% \setlength{\tabcolsep}{1pt}
% \renewcommand{\arraystretch}{1.1}
% \setlength{\abovecaptionskip}{-0.1cm} 
% \caption{Quantitative comparison on the VFI tasks on BS-ERGB, DAVIS and Pexels dataset (time $\times$ 12).}
% \begin{center}
% \begin{tabular}{l|ccccc}
% \hline
% \rowcolor{gray!10}
% \textbf{BS-ERGB}& {PSNR$\uparrow$} & {SSIM$\uparrow$} & LPIPS$\downarrow$ &    FID$\downarrow$ & FVD$\downarrow$  \\ \hline
% VDM-EVFI  &  \textbf{23.590}  & \textbf{0.741}   &  0.184    & 30.335     & 299.201   \\ 
% {Ours} &  23.437  & 0.706  & \textbf{0.131}  & \textbf{10.532}  & \textbf{146.095}  \\
% \rowcolor{gray!10}
% \textbf{DAVIS} &  {PSNR$\uparrow$} & {SSIM$\uparrow$} & LPIPS$\downarrow$ &    FID$\downarrow$ & FVD$\downarrow$  \\ \hline
% VDM-VFI   & 23.831   & 0.767  & 0.253 & 52.178  & 510.708 \\ 
% {Ours}  &   \textbf{ 25.918 } & \textbf{0.803 } & \textbf{ 0.113 }  & \textbf{ 17.138}  & \textbf{ 183.573} \\
% \hline
% \rowcolor{gray!10}
%  \textbf{Pexels}&{PSNR$\uparrow$} & {SSIM$\uparrow$} &LPIPS$\downarrow$& FID$\downarrow$& FVD$\downarrow$ \\ \hline
% VDM-VFI    &27.715  & 0.844  & 0.194 & 42.654 & 401.272 \\
% {Ours} & \textbf{ 29.411} & \textbf{0.861 } & \textbf{ 0.079} & \textbf{ 18.905} & \textbf{ 168.214} \\
% \hline
%  \end{tabular}
% \end{center}
% \label{table:comparison_result_on_x12}
% \end{table}

\begin{figure*}[th!]
  \centering
\includegraphics[width=0.8\linewidth]{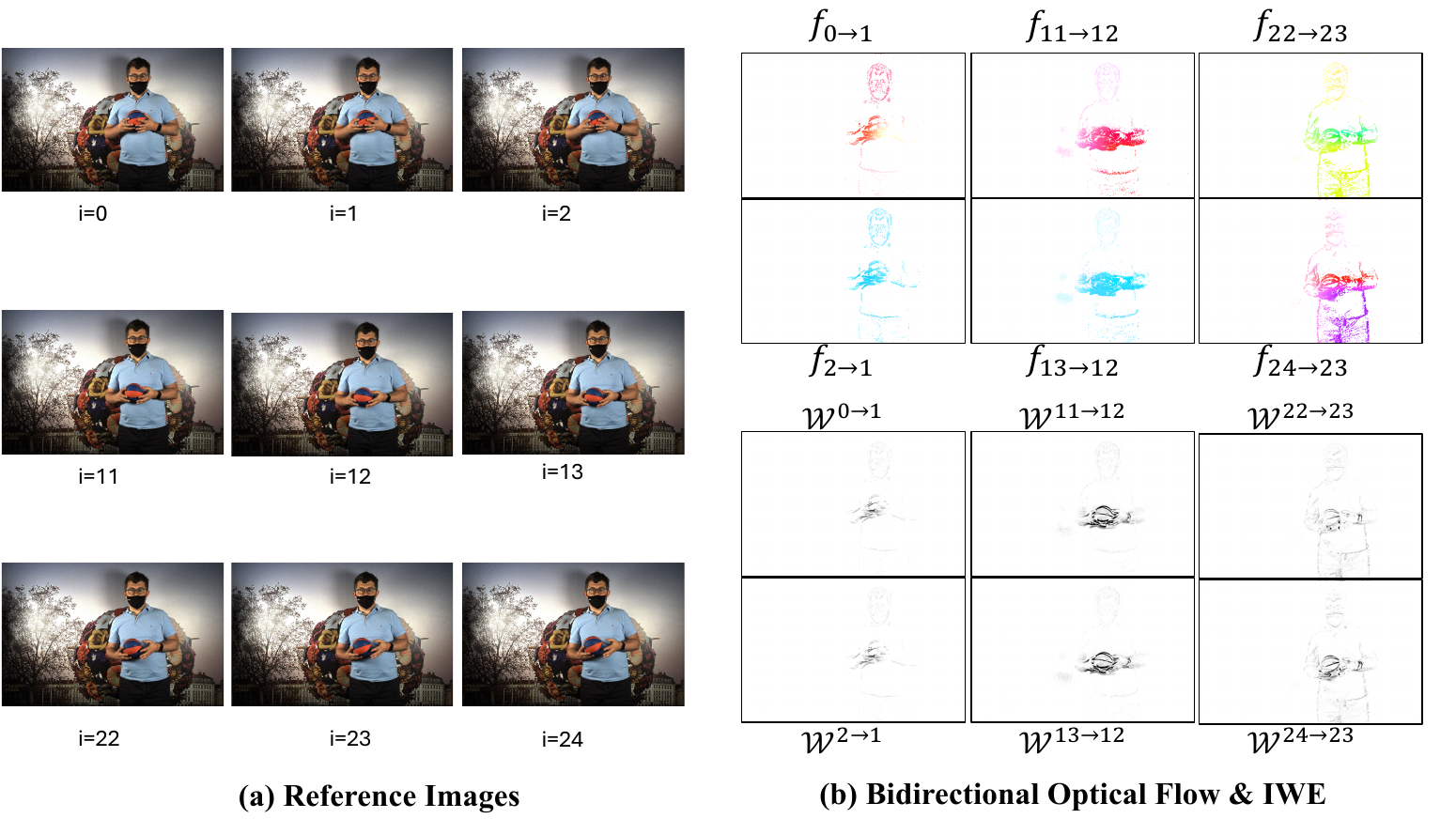}
 \vspace{-10pt}
 \caption{Illustration of the optical flow \& IWE.}
  \label{fig:opticalflow_vis}
  \vspace{-15pt}
  \end{figure*}

\subsection{Ablation Studys}
\subsubsection{Impact of Different Input Features}
To systematically evaluate the contribution of each input component to the overall interpolation performance, we conduct an ablation study by selectively removing or modifying specific modules in our framework and retraining each variant from scratch. The ablation settings include: 
\begin{enumerate}
    \item \textbf{No IWE and flows:} Only fine-tunes the Wan2.1 FLF2V model on our dataset, without incorporating any event-based features (including IWE and flows). 
    \item \textbf{w/o optical flow:} Removes the optical flow warping module while retaining the IWE encoder. 
    \item \textbf{w/o IWE:} Removes the IWE encoder while keeping the optical flow warping and fusion module. 
    \item \textbf{IWE and flow as input:} Uses both IWE and flow features as direct inputs, but disables the warping mechanism.
    \item \textbf{Full model}: Use IWE as input, and flows as warping manner to align the temporal features.

\end{enumerate}

All ablation variants are individually trained for 5,400 steps on a NVIDIA A800 GPU and evaluated on the BS-ERGB test set. The quantitative results are presented in Tab.~\ref{table:ablation_result_rebuttal}.
The results confirm that both IWE and optical flow features play essential roles in enhancing interpolation quality. Directly injecting flow information as input (without warping) yields inferior performance compared to the warping-based approach, highlighting the effectiveness of explicit temporal alignment via flow-guided feature warping.

\begin{table}[!h]
\small
\centering
\setlength{\tabcolsep}{2pt}
\renewcommand{\arraystretch}{1.2}
\setlength{\abovecaptionskip}{-1pt} 
%\vspace{-5pt}
\caption{Ablation study of the proposed components on the BS-ERGB test set (time $\times$ 24).}
\begin{center}
\begin{tabular}{c|ccccccc}
\hline
\rowcolor{gray!10}
\textbf{Methods}  &PSNR$\uparrow$& SSIM$\uparrow$ & LPIPS$\downarrow$& FID$\downarrow$ & FVD$\downarrow$ \\ \hline
 w/o (IWE and flows) & 17.390 & 0.578 & 0.241& 22.233& 678.026  \\ 
 w/o flows warping& 22.751 & 0.685 & 0.126 & 9.207 & 124.715\\ 
 w/o IWE  & 21.237& 0.626& 0.189 & 17.807& 200.861 \\
 IWE $\&$ flow inputs & 22.839 & 0.689& 0.126& 9.402& 123.306\\
Full model & \textbf{23.072}& \textbf{0.693}& \textbf{0.126} & \textbf{9.023}& \textbf{121.620}  \\
\hline
\end{tabular}
\end{center}
\label{table:ablation_result_rebuttal}
\vspace{-5pt}
\end{table}

\subsubsection{Impact of Different Injection Blocks}
In this work, the Wan2.1 FLF2V backbone contains 40 blocks. We choose the first two blocks as the injection points for our flow-based alignment and fusion module. To further investigate the effect of injection position, we conduct additional ablation studies by inserting the module into the middle two blocks and the last two blocks of the backbone. The corresponding results are reported in Tab.~\ref{table:ablation_result_block_position_rebuttal}.
The results indicate that injecting temporal information into the last two blocks yields better LPIPS, FID, and FVD scores, whereas injecting it into the first two blocks leads to better PSNR and SSIM. In this work, we adopt the first two blocks as the injection position to prioritize reconstruction quality.

\subsubsection{Impact of Different Event Representation}
We conducted a literature review on event-based conditioning and identified two representative approaches: the edge-based conditioning method proposed in CUBE~\cite{zhao2024controllable}, which converts the event stream into edge images, and the event voxel stack representation used in VDM-EVFI~\cite{chen2024repurposing}.
We incorporated each of these event representations separately as additional inputs to the Wan2.1 FLF2V by feeding them through an encoder that shares the same architecture as our IWE encoder. The models were trained for 5,400 steps on an NVIDIA A800 GPU and evaluated on the BS-ERGB dataset for fair comparison. The results are summarized in Tab.~\ref{table:ablation_event_rep_rebuttal}.  These results demonstrate that our event representation (IWE and Flow Warping) achieves consistently superior performance across all metrics under the same backbone architecture and training settings.

\begin{table}[!h]
\small
\centering
\setlength{\tabcolsep}{2pt}
\renewcommand{\arraystretch}{1.2}
\setlength{\abovecaptionskip}{-1pt} 
%\vspace{-5pt}
\caption{Ablation study on BS-ERGB test dataset regarding the injection position of flow-guided blocks (time ×24).}
\begin{center}
\begin{tabular}{c|ccccccc}
\hline
\rowcolor{gray!10}
\textbf{ Injection Point}  &PSNR$\uparrow$& SSIM$\uparrow$ & LPIPS$\downarrow$& FID$\downarrow$ & FVD$\downarrow$ \\ \hline
 Last two blocks& 22.814 & 0.691 & \textbf{0.120} & \textbf{8.986} & \textbf{108.093}   \\ 
 Mid. two blocks & 22.801 & 0.687& 0.129& 9.393& 121.771\\ 
First two blocks & \textbf{23.072}& \textbf{0.693}& 0.126 & 9.023& 121.620  \\
\hline
\end{tabular}
\end{center}
\vspace{-10pt}
\label{table:ablation_result_block_position_rebuttal}
%\vspace{-15pt}
\end{table}

\begin{table}[!h]
\small
\centering
\setlength{\tabcolsep}{2pt}
\renewcommand{\arraystretch}{1.2}
\setlength{\abovecaptionskip}{-1pt} 
%\vspace{-5pt}
\caption{Ablation study on BS-ERGB test dataset of different event conditions (time $\times$ 24).}
\begin{center}
\begin{tabular}{c|ccccccc}
\hline
\rowcolor{gray!10}
\textbf{Event Rep.}  &PSNR$\uparrow$& SSIM$\uparrow$ & LPIPS$\downarrow$& FID$\downarrow$ & FVD$\downarrow$ \\ \hline
Edge  &17.703  & 0.599 & 0.227  & 23.204 &755.380  \\
 Event Voxel Stack&21.311& 0.611& 0.205& 18.447&228.950\\
Ours (IWE $\&$ Flow) & \textbf{23.072}& \textbf{0.693}& \textbf{0.126} & \textbf{9.023}& \textbf{121.620}  \\
\hline
\end{tabular}
\end{center}
\vspace{-12pt}
\label{table:ablation_event_rep_rebuttal}
%\vspace{-15pt}
\end{table}

\subsection{Visualization of IWEs and Optical Flows}
We provide a detailed visualization of the sparse bidirectional optical flow and IWE, as shown in Fig.~\ref{fig:opticalflow_vis} (b). Specifically, we apply a contrast maximization method on the event data between every two consecutive latent frames to compute these sparse optical flow segments and the IWE. Comparing them with the ground truth in Fig.~\ref{fig:opticalflow_vis} (a) reveals that the optical flow segments in Fig.~\ref{fig:opticalflow_vis} (b) accurately capture the motion between consecutive frames in Fig.~\ref{fig:opticalflow_vis} (a).

\clearpage
\section{Conclusion}
In this paper, we explore leveraging event data to efficiently enhance DiT-based video frame interpolation tasks. We propose an adapter-based framework that integrates high temporal resolution cues from event cameras—capturing continuous motion data via a pre-trained Image-to-Video (I2V) model, requiring only lightweight adapter training. By incorporating IWE and bidirectional sparse optical flow, our approach enables precise temporal guidance, mitigating motion artifacts and improving interpolation quality.
Our experimental results demonstrate that event-enhanced interpolation outperforms existing methods in terms of both accuracy and temporal consistency, effectively reducing long-range motion drift and improving structural fidelity. This confirms the feasibility of extracting optical flow and IWE from event data to assist frame interpolation, thereby circumventing the challenges associated with directly adapting sparse event data to dense RGB frames.

\bibliographystyle{ACM-Reference-Format}
\bibliography{sample-base}


\end{document}